\documentclass[11pt,a4paper]{article}
\usepackage[hyperref]{acl2020}
\usepackage{times}
\usepackage{latexsym}

\usepackage{microtype}

\aclfinalcopy % Uncomment this line for the final submission
\usepackage{amsmath}
\usepackage{amssymb}
\usepackage{graphicx}

\usepackage{framed}
\usepackage{subfigure}
\usepackage{booktabs}
\usepackage{empheq}
\usepackage{array}
\usepackage{color,soul}
\usepackage{enumitem}
\usepackage{multirow}
\usepackage{mathtools}

\newcommand{\rrr}[2]{\small{#1$ \pm $#2}}

\DeclarePairedDelimiterX{\infdivx}[2]{(}{)}{%
	#1\;\delimsize\|\;#2%
}

\newcommand{\ourDataName}{\textsc{Psg2Sum}}

\title{Learning to Summarize Passages: Mining Passage-Summary Pairs from Wikipedia Revision Histories}

\author{Qingyu Zhou$^\dag$\thanks{\; Contribution during internship at Microsoft Research.} \hspace{0.15cm} Furu Wei$^\ddag$ \hspace{0.15cm}  Ming Zhou$^\ddag$ \\
	$^\dag$Harbin Institute of Technology, Harbin, China \\
	$^\ddag$Microsoft Research, Beijing, China \\
	{\tt qyzhgm@gmail.com} \hspace{0.25cm} {\tt \{fuwei,mingzhou\}@microsoft.com}
}

\date{}

\begin{document}
\maketitle
\begin{abstract}
	%The development of automatic summarization systems is constrained by the limited paired training data.
	In this paper, we propose a method for automatically constructing a passage-to-summary dataset by mining the Wikipedia page revision histories.
	%In particular, the method mines the article passages and the introduction sentences which are added to the pages simultaneously.
	In particular, the method mines the main body passages  and the introduction sentences which are added to the pages simultaneously.
	The constructed dataset contains more than one hundred thousand passage-summary pairs.
	The quality analysis shows that it is promising that the dataset can be used as a training and validation set for passage summarization.
	We validate and analyze the performance of various summarization systems on the proposed dataset.
	The dataset will be available online at \url{https://res.qyzhou.me/}.
	
\end{abstract}

\section{Introduction}
The area of automatic text summarization has received a lot of attention recently~\cite{rush-chopra-weston:2015:EMNLP,cheng-lapata:2016:P16-1,nallapati2017summarunner,see-liu-manning:2017:Long,zhou-EtAl:2017:Long,tan2017abstractive}.
Many recent summarization models are working on two types of input, i.e., sentence level summarization~\cite{rush-chopra-weston:2015:EMNLP,chopra-auli-rush:2016:N16-1, nallapatiabstractive,zhou-EtAl:2017:Long} and single document level  summarization~\cite{cheng-lapata:2016:P16-1,see-liu-manning:2017:Long,nallapati2017summarunner}.
The development of these neural network summarization systems requires relatively large datasets~\cite{rush-chopra-weston:2015:EMNLP,hermann2015teaching}.

The sentence level summarization dataset is constructed automatically by pairing the title and the first sentence in a news article~\cite{rush-chopra-weston:2015:EMNLP}.
The input sentence and the output title are extracted and cleaned heuristically from Annotated English Gigawords~\cite{Napoles:2012:AG:2391200.2391218}.
The  document level datasets being used frequently are newswire datasets such as CNN, Daily Mail and NY Times, which are usually used to produce several sentences as the summary.
However, to the best of our knowledge, no prior work has discussed summarizing a text \textit{passage} which has the potential use for long document summarization, slides highlight generation~\cite{wang2017phrase}, language teaching~\cite{huang2015study} and so on.
The above-mentioned datasets are either for sentence or document summarization, which ignores the passage granularity.

In this paper, we introduce a new summarization dataset which aims to explore the \textbf{passage}-to-summary granularity of text summarization task.
We make the key observation that in two temporally adjacent Wikipedia page revisions, the passage in the article body and the sentence in the introduction, which are added simultaneously to a Wikipedia page, are possibly a passage-summary pair.
Based on this assumption, we mine the English Wikipedia history dump to extract possible pairs.
By cleaning and filtering the extracted data, we created a new \textbf{p}a\textbf{s}sa\textbf{g}e-to-\textbf{sum}mary (\textbf{\ourDataName{}}) dataset which contains 100,118 examples.
Quality analysis and the comparison to other summarization datasets show that it is promising that \ourDataName{} can be used as a training and evaluation dataset.

\begin{table}[ht]
	\small
	\begin{center}
		\begin{tabular}{p{0.45\textwidth}}
			\hline
			The \hl{collision} between \hl{trains 608} and \hl{653} happened on kilometer 8.055 at 17:42 (some sources says at 17:44). The speed of the steam train 608 was about \hl{55 km/h}, train 653 about \hl{60 km/h}. Both drivers tried to slow in the loose , but it was too late.\\
			\hline
			A passenger steam train 608 at speed 55 km/h abreast collided with a diesel railcar 653 at speed 60 km/h. \\
			\hline
		\end{tabular}
	\end{center}
\caption{\label{tbl:intro_example}A passage-to-summary example in the \ourDataName{} dataset. The passage (top) in the article and the sentence (bottom) in the lead section were added to the Wikipedia page simultaneously. The key information in the passage is highlighted.}
\end{table}

Our primary contributions are:
\begin{itemize}
	\item A scalable, language agnostic method to create a passage-to-summary dataset from Wikipedia revision history.
	\item Fill the granularity vacancy of summarization datasets that we first present an \textit{open-domain}, \textbf{\textit{passage}}-to-summary corpus.
	\item Publicly release of the English \ourDataName{} dataset on an anonymous URL for double-blind review.
	\item The English version of \ourDataName{} dataset will be available online at  \url{https://res.qyzhou.me/}.
	\item We validate the performance of various summarization methods on \ourDataName{}.
\end{itemize}
\section{The \ourDataName{} Dataset}
%Existing summarization datasets
%\subsection{Mining Passage-Summary Pairs from Wikipedia History}
\subsection{Dataset Creation}

Wikipedia maintains the history of its pages which contains a list of the pages' previous revisions\footnote{\footnotesize\url{https://en.wikipedia.org/wiki/Help:Page_history}}.
The page revisions have been exploited for some NLP tasks, such as sentence splitting~\cite{D18-1080}, sentence compression~\cite{yamangil2008mining} and sentence simplification~\cite{woodsend2011learning,yatskar2010sake}

Most of the Wikipedia articles have \textbf{lead section}s\footnote{\url{https://en.wikipedia.org/wiki/Wikipedia:Manual_of_Style/Lead_section}} (also known as the lead or introduction, screenshot available in the Appendix).
It serves as an introduction to the article and a summary of its most important contents.
Therefore, we pair the passages in the main body and the sentences in the lead section to construct the \ourDataName{} corpus.
We make the assumption that in a page revision, a sentence added to the lead section is possibly the summary of one passage added to the article at the same time.
Based on this assumption, we compare two temporally adjacent revisions of a page to extract the additions.

We first extract and clean text by stripping the Wikipedia markup language (wikicode) using \texttt{mwparserfromhell}\footnote{\url{https://github.com/earwig/mwparserfromhell}}.
Then the text in the lead section is split into sentences using the sentence splitting algorithm in Moses~\cite{koehn2007moses}\footnote{We use a python implementation: \url{https://github.com/berkmancenter/mediacloud-sentence-splitter}}.
The sentences are then tokenized with the spaCy tokenizer.
We compare the processed two page revisions using Python's \texttt{difflib} to extract the added sentences and passages.

Given all the added sentences $ S = \lbrace s_{1}, \dots ,s_{n} \rbrace $ in the lead section and the passages $ P = \lbrace  p_{1}, \dots ,p_{n} \rbrace $ in the article,  we use some heuristics to mine passage-summary pairs from them.
Similar to \citet{rush-chopra-weston:2015:EMNLP}, we find the possible candidates by calculating the unigram overlap to ensure the passage-summary relationship.
Specifically, for the candidate passage-summary pair $ (p_{j} ,s_{i} ) $, we first remove the stopwords from both the sentence $ i$ and passage $ j $ to get $ (p_{j}', s_{i}') $
The candidate score is defined as the overlap rate:
\begin{equation}
score\left((p_{j} ,s_{i})\right) = \frac{\lvert \lbrace w \vert w \in s_{i}', w \in p_{j}' \rbrace \rvert}{\lvert s_{i}' \rvert}
\end{equation}

For the candidate sentence $ s_{i} $ in the lead section, we choose the passage with the maximum overlap score $ score\left((p_{j} ,s_{i})\right) $.
To filter out the misaligned passage-summary pairs, we set a minimum overlap rate threshold $ \lambda $.
Specifically, if $ score\left((p_{j} ,s_{i})\right) $ is less than $ \lambda $, we discard the candidate pair $ (p_{j} ,s_{i}) $.

\begin{table*}[h]
	\small
	\begin{center}
		\begin{tabular}{lcccccccc}
			%		\begin{tabular}{@{~}l@{\hspace{1ex}}c@{\hspace{1ex}}c@{\hspace{1ex}}c@{~}}
			\toprule
			\bf \multirow{2}{*}{\bf Dataset} & \multirow{2}{*}{\bf Granularity} & \multirow{2}{*}{\bf Domain}  & \multirow{2}{*}{\bf Corpus Size}   & \multicolumn{2}{c}{\bf avg. Input Length}  &  \multicolumn{2}{c}{\bf Output Length} & \bf Reference  \\ 
			 &  &  & & sentences & words & sentences & words & \bf Number \\
			\midrule
			DUC2002 (task 1) & Doc & News & 567  & 27.37 & 629.64 & 10.19 & 215.09 & 1.96 \\
			Gigawords & Sentence & News & 3.8m & 1 & 31.35 & 1 & 8.23 & 1 \\
			%			msr-atc \red{cite} & Sentence  &  & & & \\
			CNN & Doc & News & 92,579 & 33.98 & 760.50 & 3.59 & 45.70 & 1 \\
			Daily Mail & Doc & News&  219,506 & 29.33 & 653.33 & 3.86 & 54.65 & 1 \\
			NY Times & Doc & News& 654,759 & 35.55 & 800.04 & 2.44 & 45.54 & 1 \\
			\hline
			\ourDataName{} & \textbf{Passage} & \textbf{Open} & 100,118 & \textbf{4.83} & \textbf{118.26} & 1 & 22.20 & 1 \\
			\bottomrule
		\end{tabular}
	\end{center}
	\caption{\label{tbl:compare}A comparison of current summarization datasets and \ourDataName{}.}
\end{table*}

\subsection{Quality and Statistics of  \ourDataName{}}
As the heuristic method cannot guarantee all the pairs are true passage-summary pairs, we manually check the quality of the constructed dataset.
We randomly sample 50 examples and label them as the following:
\begin{itemize}
	\item Good: The sentence is a summary of the given passage.
	\item Unsupported: The sentence is irrelevant to the passage. Or, some important content cannot be found in the passage, such as dates and places, which makes it not understandable.
\end{itemize}
Furthermore, we do the same labeling on 50 random examples from the English Gigawords sentence summarization dataset~\cite{rush-chopra-weston:2015:EMNLP} which is also created automatically and cleaned with heuristics.

%\begin{table}[h]
%%	\small
%	\begin{center}
%%		\begin{tabular}{c|ccc|c}
%		\begin{tabular}{@{~}c|@{\hspace{1ex}}c@{\hspace{1ex}}c@{\hspace{1ex}}c@{~}|c@{\hspace{1ex}}}
%			%		\begin{tabular}{@{~}l@{\hspace{1ex}}c@{\hspace{1ex}}c@{\hspace{1ex}}c@{~}}
%			\hline
%			\textit{Thresh.} $ \lambda $  & Good &  Unsup. & Good Rate &  Size\\
%			\hline
%			0.5 & 27 & 23 & 54\% & 117,026 \\
%			0.6 & 33 & 17 & 66\% & 100,118 \\
%			0.7 & 34  & 16 & 68\% & 68,070 \\
%			
%			\hline
%			Gigawords  & 28 & 22 & 56\% & 3.8m  \\
%			\hline
%		\end{tabular}
%	\end{center}
%	\caption{\label{tbl:thresh}Quality vs corpus size trade-off when setting the minimum overlap \textbf{thresh}old value $ \lambda $. The \textbf{Good} and \textbf{Unsup}ported numbers are counted in a 50 random sampled subset.}
%\end{table}

\begin{table}[h]
	%	\small
	\begin{center}
				\begin{tabular}{c|cc|c}
%		\begin{tabular}{@{~}c|@{\hspace{1ex}}c@{\hspace{1ex}}c@{\hspace{1ex}}c@{~}|c@{\hspace{1ex}}}
			%		\begin{tabular}{@{~}l@{\hspace{1ex}}c@{\hspace{1ex}}c@{\hspace{1ex}}c@{~}}
			\hline
			\textit{Thresh.} $ \lambda $  & Good (\%) &  Unsup. &   Size\\
			\hline
			0.5 & 27 (54\%) & 23 &   117,026 \\
			0.6 & 33 (66\%) & 17 &   100,118 \\
			0.7 & 34 (68\%)  & 16 &   68,070 \\
			
			\hline
			Gigawords  & 28 (56\%) & 22 &   3.8m  \\
			\hline
		\end{tabular}
	\end{center}
	\caption{\label{tbl:thresh}Quality vs corpus size trade-off when setting the minimum overlap \textbf{thresh}old value $ \lambda $. The \textbf{Good} and \textbf{Unsup}ported numbers are counted in a 50 random sampled subset.}
\end{table}

As shown in Table~\ref{tbl:thresh}, overlap threshold $ \lambda = 0.6 $ is a good trade-off between the Good rate and the corpus size.
For the 50 random examples, increase the threshold $ \lambda $ from 0.5 to 0.6 leads to a 12\% absolute Good rate improvement with only 6,808 examples filtered.
When increasing the threshold $ \lambda $ from 0.6 to 0.7, we only observe 2\% Good rate improvement but the corpus size drastically shrinks to 68,070.
Compared to the 56\% good rate of the successful English Gigawords dataset, we choose the threshold value $ \lambda = 0.6 $.

After filtering and cleaning, the final \ourDataName{} dataset contains 100,118 passage-summary pairs.
We randomly split the dataset into training, validation and testing sets, which have 92,118, 4000 and 4000 passage-summary pairs respectively.

\subsection{Comparison to Other Datasets}
Since 2001, NIST had organized the DUC summarization tasks~\cite{over2007duc}.
They provided high-quality, human-created document/multi-document summarization datasets.
However, DUC dataset is too small to train an abstractive summarization system using artificial neural networks.
For example, DUC 2002 task 1 only contains 567 documents associated with around 1.96 references.
%Therefore, including \ourDataName{}, recent large scale datasets are more suitable for training deep neural networks to build abstractive summarization systems.
Therefore, large scale datasets are necessary for training neural abstractive summarization systems.

Abstractive sentence summarization has attracted research focus in recent years~\cite{rush-chopra-weston:2015:EMNLP,toutanova-EtAl:2016:EMNLP2016,chopra-auli-rush:2016:N16-1,nallapatiabstractive}.
\citet{rush-chopra-weston:2015:EMNLP} propose constructing a sentence summarization (or headline generation) dataset by pairing the first sentence and the title in a news article.
They use the Annotated English Gigawords~\cite{Napoles:2012:AG:2391200.2391218} as the article source.
As shown in Table~\ref{tbl:thresh}, though the Gigawords corpus contains some noise, it is still useful as a training and evaluation dataset.
Considering the Good rate of \ourDataName{} is about 10\% higher than the English Gigawords dataset, it is promising that \ourDataName{} can achieve the same goal.

Recently, newswire websites such as CNN, Daily Mail and NY Times have been used as sources for single document summarization.
The NY Times is currently the largest summarization dataset as shown in Table~\ref{tbl:compare}.
However, it is bias toward extractive strategies, and limited work has used this dataset for summarization~\cite{N18-1065}.
CNN and Daily Mail~\cite{hermann2015teaching} have been frequently used in recent document summarization research.
These datasets have been used for summarization as is \cite{see-liu-manning:2017:Long}, or after pre-processing for entity anonymization \cite{nallapati2017summarunner}.
Additionally, some systems mix CNN and Daily Mail as training data \cite{nallapati2017summarunner,see-liu-manning:2017:Long,paulus2017deep}, whereas others use only Daily Mail articles~\cite{cheng-lapata:2016:P16-1,nallapati2016classify}.
%These different usages in previous works make comparisons between systems using these data challenging.
Therefore, it would be challenging for systems to make comparisons considering that previous works are using different versions of datasets.
%Moreover, \citet{N18-1065} point out that CNN/Daily Mail is biased toward extractive summarization.
%Moreover, as shown in Table~\ref{tbl:compare}, these news documents contain 

\begin{table*}[ht]
	%		\small
	\setlength{\tabcolsep}{0.9pt}
	\begin{center}
		\begin{tabular}{l|c|c|c|c|c|c|c|c|c}
			\hline
			\multirow{2}{*}{\bf Models} & \multicolumn{3}{c|}{\textsc{Rouge}-1} & \multicolumn{3}{c|}{\textsc{Rouge}-2} & \multicolumn{3}{c}{\textsc{Rouge}-L} \\ \cline{2-10}
			& R & P & F1 & R & P & F1  & R & P & F1 \\
			\hline
			\small s2s & \rrr{36.31}{0.66} &  \rrr{35.24}{0.71} & \rrr{33.35}{0.60} & \rrr{18.08}{0.66} & \rrr{17.95}{0.68} &  \rrr{16.78}{0.60} &  \rrr{31.57}{0.66} & \rrr{30.64}{0.70} & \rrr{29.00}{0.60} \\
%			\small XFMR & \rrr{26.22}{0.52} & \rrr{35.38}{0.68} & \rrr{28.57}{0.52} & \rrr{9.23}{0.44} & \rrr{12.38}{0.52} & \rrr{10.02}{0.42}  & \rrr{22.70}{0.50} & \rrr{30.25}{0.63} & \rrr{24.61}{0.50} \\
			\hline
			\small s2s+copy & \rrr{35.51}{0.73} & \textbf{\rrr{37.38}{0.79}} & \rrr{33.78}{0.66} & \rrr{18.68}{0.67} & \textbf{\rrr{19.87}{0.76}} &  \rrr{17.80}{0.65}  & \rrr{30.84}{0.69 } & \textbf{\rrr{32.64}{0.75}} & \rrr{29.43}{0.65} \\
			\small PGN &  \rrr{36.27}{0.76} &  \rrr{36.57}{0.77} & \rrr{33.99}{0.66} & \rrr{19.05}{0.72} & \rrr{19.56}{0.75} & \rrr{17.99}{0.67}  & \rrr{31.34}{0.73} & \rrr{31.82}{0.73} & \rrr{29.49}{0.64} \\
			\hline
			\small LEAD1 & \rrr{42.97}{0.77} & \rrr{35.35}{0.73} & \rrr{35.97}{0.66} & \rrr{22.76}{0.72} & \rrr{18.71}{0.71} & \rrr{19.01}{0.65}  & \rrr{36.06}{0.73} & \rrr{29.88}{0.70} & \rrr{30.29}{0.63} \\
			\small TextRank & \rrr{39.95}{0.77} & \rrr{33.56}{0.73} & \rrr{33.74}{0.64} & \rrr{20.01}{0.75} & \rrr{16.95}{0.71} & \rrr{16.92}{0.65}  & \rrr{33.29}{0.73} & \rrr{28.13}{0.69} & \rrr{28.18}{0.62} \\
			\small NN-SE & \textbf{\rrr{43.76}{0.80}} & \rrr{35.11}{0.74} & \textbf{\rrr{36.21}{0.67}} & \textbf{\rrr{23.19}{0.75}} & \rrr{19.14}{0.73} & \textbf{\rrr{19.40}{0.68}}  & \textbf{\rrr{36.59}{0.74}} & \rrr{29.61}{0.73} & \textbf{\rrr{30.40}{0.63}} \\
			\hline
		\end{tabular}
	\end{center}
	\caption{\label{tbl:result}\textsc{Rouge} evaluation results on \ourDataName{} of various summarization models. The scores with 95\% confidence interval are given by the official \textsc{Rouge} script. The best results are in \textbf{bold}.}
\end{table*}

All the above-mentioned datasets, including both the sentence level and the document level summarization datasets, are constructed or labeled using the newswire source, which leads to the fact that they are all in the \textit{news domain}.
The proposed \ourDataName{} is constructed with the open-domain Wikipedia~\cite{chen2017reading,yang2015wikiqa}.
As far as we know, this is the first open-domain text summarization dataset.
Table~\ref{tbl:compare} summarizes the key features of existing summarization datasets and \ourDataName{}.
To the best of our knowledge, \ourDataName{} is the first \textbf{\textit{passage}}-to-summary dataset, which is with the same magnitude with the current frequently used CNN and Daily Mail datasets.
The average input length of \ourDataName{} is 4.83 sentences (118.26 words), compared with the average length 33.98 sentences (760.50 words) of CNN and 29.33 sentences (653.33 words) of  Daily Mail corpus.

\section{Experiments}

\subsection{Models}
%We evaluate several summary methods, including abstractive and extractive, supervised and unsupervised methods, on the \ourDataName{} dataset:
We evaluate several summary models on the \ourDataName{} dataset and the detailed model configurations can be found in the Appendix:
\begin{description}
	\item[s2s] (sequence-to-sequence) is a basic neural text generation model proposed by~\newcite{sutskever2014sequence}.
	In this work, we use the RNN-based s2s model with attention mechanism~\cite{bahdanau2014neural}.
	\item[s2s+copy] is an extension of s2s incorporated with copying mechanism~\cite{gu-EtAl:2016:P16-1,gulcehre-EtAl:2016:P16-1}.
	\item[PNG] (Pointer-Generator Network) \cite{see-liu-manning:2017:Long} is an extension of s2s with copying and coverage~\cite{P16-1008} mechanisms.
%	\item[Transformer] (XFMR) is with the encoder-decoder architecture which are built with Multi-Head Attention~\cite{vaswani2017attention}.
	\item[LEAD1] extracts the first sentence as the summary. The leading sentences baseline is also a strong baseline on newswire datasets such as CNN, Daily Mail and NY Times.
	\item[TextRank]~\cite{mihalcea2004textrank} is an unsupervised extractive method.
	We use the implementation in the Gensim package~\cite{rehurek_lrec}.
	\item[NN-SE]~\cite{cheng-lapata:2016:P16-1} is an extractive neural model with a hierarchy architecture.
	It predicts the probability of being extracted for each sentence.
\end{description}

\subsection{Evaluation Metric}
We use \textsc{Rouge} (version 1.5.5)~\citep{lin2004rouge} as our evaluation metric.
\textsc{Rouge} measures the quality of summary by computing overlapping lexical units, such as unigram, bigram, trigram, and longest common subsequence (LCS).
Following previous works, we report \textsc{Rouge}-1, \textsc{Rouge}-2 and \textsc{Rouge}-L metrics in the experiments.

\subsection{Results}
We validate various models on the \ourDataName{} dataset, including abstractive models (s2s), extractive models (LEAD1, TextRank, NN-SE) and mixed models (s2s+copy, PGN).
Table~\ref{tbl:result} shows the \textsc{Rouge} evaluation results.
We observe that extractive methods perform better in terms of \textsc{Rouge} Recall.
For example, the NN-SE model achieves the best recall performance among all the baseline models, i.e., 43.76 \textsc{Rouge}-1 recall and 23.19 \textsc{Rouge}-2 recall.
In the meanwhile, the abstractive models achieve better \textsc{Rouge} Precision scores.
The s2s + copy model has the best precision performance in \textsc{Rouge}-1, -2 and -L.
Surprisingly, we find that using coverage mechanism (PGN) leads to the precision drop but higher recall score (with longer output), although s2s+copy and PGN are statistically indistinguishable in terms of \textsc{F1} score.

\section{Conclusion}
In this paper, we present a heuristic approach to automatic constructing a passage-to-summary dataset, \ourDataName{}, by mining the Wikipedia page revision histories.
The quality analysis shows that it is capable of being a training and evaluation corpus despite the imperfection that it contains some noise.
Experiments on \ourDataName{} show that extractive models tend to select longer sentences and achieves higher recall score, comparing with the abstractive and mixed models' tendency to generate high precision outputs.

\bibliography{acl2020}
\bibliographystyle{acl_natbib}

\appendix

\section{Lead Section of Wikipedia}
\begin{figure*}[htbp]
	\centering
	\includegraphics[width=\textwidth]{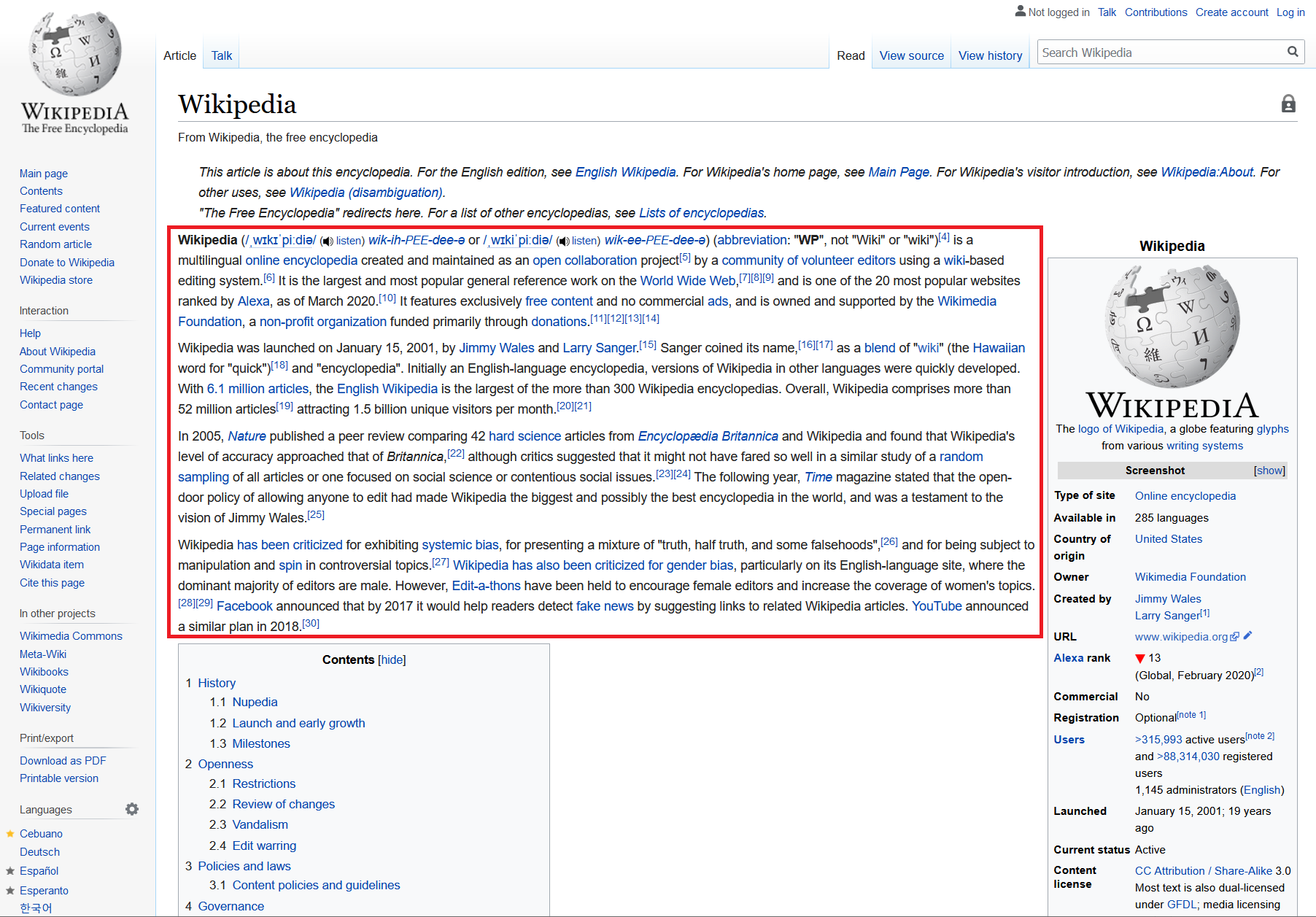}
	\caption{\label{fig:wiki} A screenshot example of the lead section of a Wikipedia article about Wikipedia.}
\end{figure*}

Figure~\ref{fig:wiki} shows a screenshot example of the lead section of a Wikipedia article about Wikipedia.

\section{Model Configurations}
\subsection{s2s}
We use the model architecture introduced in \citet{bahdanau2014neural}.
The encoder and decoder are built with Gated Recurrent Units (GRU)~\cite{cho-EtAl:2014:EMNLP2014}.
The encoder is bidirectional, with 256 dimensional forward and 256 dimensional backward backward GRU.
The decoder's hidden size is 512.
The word vector size of encoder and decoder is 300.
We use dropout~\cite{srivastava2014dropout} rate 0.5 to prevent model overfitting.
During training, we use the Adam~\cite{kingma2014adam} optimizer to learn the model with its default hyper-parameters.
The mini-batch size is set to 64.
During testing, we use beam search and the beam size is set to 5.

\subsection{s2s+copy}
The s2s+copy model is based on s2s, and augmented with copying mechanism~\cite{gu-EtAl:2016:P16-1,gulcehre-EtAl:2016:P16-1}.
The training and testing configurations are identical to the s2s model.

\subsection{PGN}
We implement Pointer-Generator Networks (PGN)~\cite{see-liu-manning:2017:Long} based on s2s+copy model by adding the coverage loss function following~\cite{see-liu-manning:2017:Long}.
The other configurations are identical to the s2s+copy model.

%\subsection{Transformer (XFMR)}
%We train a Transformer~\cite{vaswani2017attention}~ base model using the \textsc{Psg2Sum} dataset using \texttt{fairseq}\footnote{https://github.com/pytorch/fairseq/}.

\subsection{TextRank}
We use the open-source implementation of TextRank in the Gensim~\cite{rehurek_lrec} toolkit.
It refuse to summarization passages with less than three sentences.
Therefore, we randomly select one sentence as the summary for passages shorter than three sentences.

\subsection{NN-SE}
We implement NN-SE model as mentioned in the paper~\cite{cheng-lapata:2016:P16-1}.
During testing, we select the sentence with highest extraction score as the passage summary.

\section{\textsc{Psg2Sum} Data Samples}
Table~\ref{tbl:data_example} shows 5 random examples in the \textsc{Psg2Sum} dataset.

\begin{table*}[htbp]
	\centering
	\small
	\begin{tabular}{l|p{0.85\textwidth}}
		\hline
		\textbf{Example 1} & \\
		\hline
		PSG  &  A recording of the musical with 19 tracks was issued in the U.S. on Scepter Records in 1971 . It was a reissue of the 1969 Decca UK album , capitalizing on the success of 1970 's Jesus Christ Superstar in the U.S. It featured David Daltrey as Joseph , Tim Rice as Pharaoh , Dr. William S. Lloyd Webber on the Hammond organ , Alan Doggett conducting , various solo vocalists and instrumentalists , and the Colet Court choir as the chorus."Joseph And The Amazing Technicolor Dreamcoat Listing , Scepter Records , SPS-588X , 1971 " discogs.com , accessed March 17 , 2011Q\&A regarding the original Decca and Scepter albums\\
		\hline
		SUM  &  Joseph and the Amazing Technicolor Dreamcoat is a musical with lyrics by Tim Rice and music by Andrew Lloyd Webber .\\
		\hline
		
		\textbf{Example 2} & \\
		\hline
		PSG  &  In 1994 , Bush took a leave of absence from the Rangers to run for Governor of Texas against the popular incumbent , Democrat Ann Richards . On November 8 , 1994 , he defeated Richards , 53 \% to 46 \% . As Governor , Bush forged a legislative alliance with powerful Texas Lt . Governor Bob Bullock , a longtime Democrat . In 1998 Bush went on to win re - election in a landslide victory with nearly 69 \% of the vote , becoming the first Texas governor to be elected for two consecutive four - year terms .   During Bush 's governorship , he undertook significant legislative changes in criminal justice , tort law , and school financing . Bush took a hard line on capital punishment and received much criticism from advocates wanting to abolish the death penalty . Under Bush , Texas ' incarceration rate was 1014 inmates per 100,000 state population in 1999 , the second highest in the nation , owing mainly to strict enforcement of drug laws . In September 1999 , Bush signed the Texas Futile Care Law . Bush 's transformative agenda and family pedigree now provided an opportunity to advance his political career to the national level .\\
		\hline
		SUM  &  Bush was elected 46th Governor of Texas in 1994 and re - elected in 1998 .\\
		\hline
		
		\textbf{Example 3} & \\
		\hline
		PSG  &  The group 's first single , " Saturday Night Party ( Read My Lips ) " , was an immediate success , and became an Ibiza anthem during the summer of 1993 . It became their first Top 40 hit in the United Kingdom , peaking at \# 29 . After introducing a singer to the group ( Shanie Campbell ) , they released the single " Do n't Give Me Your Life " in 1994 , being an extended remix to the original " Alex Party " track . It reached \# 2 in both Ireland and the United Kingdom ( their highest charting hit in those countries ) and \# 13 in Australia , plus it topped the Club Record category at Music Week 's 1995 Awards . It was included in many compilation albums all over the world , and remains their most famous release .\\
		\hline
		SUM  &  Their most famous single to date is " Do n't Give Me Your Life " , a \# 2 hit in both Ireland and the United Kingdom in early 1995.\\
		\hline
		
		\textbf{Example 4} & \\
		\hline
		PSG  &  Throughout the existence of medieval Livonia there was a constant struggle for superiority in the rule over the lands by the Church , the order , the secular nobles of German descent who ruled the fiefs and the citizens of the Hanseatic town of Riga . Two major civil wars were fought in 1296 - 1330 , 1313 - 1330 , and in 1343 - 1345 the Estonian revolt resulted in the annexation of the Danish Duchy of Estonia within the Teutonic Ordensstaat .\\
		\hline
		SUM  &  Throughout the existence of medieval Livonia there was a constant struggle over the supremacy of ruling the lands by the Church , the Order , the secular German nobility and the citizens of the Hanseatic towns of Riga and Reval .\\
		\hline
		
		\textbf{Example 5} & \\
		\hline
		PSG  &  Along with Matsumoto Castle and Kumamoto Castle , Himeji Castle is considered one of Japan 's three premier castles . It is the most visited castle in Japan , receiving over 820,000 visitors annually . Starting in April 2010 , Himeji Castle underwent restoration work to preserve the castle buildings , and reopened to the public on 27 March 2015 . \\
		\hline
		SUM  &  In order to preserve the castle buildings , it underwent restoration work for several years and reopened to the public on March 27 , 2015 .\\
		\hline
		
	\end{tabular}
	\caption{\label{tbl:data_example}5 random examples from the \textsc{Psg2Sum} dataset.}
\end{table*}

\end{document}